\pdfoutput=1
\documentclass[11pt,a4paper]{article}
\usepackage[hyperref]{emnlp2020}
\usepackage{times}
\usepackage{latexsym}

\usepackage{microtype}

\aclfinalcopy 

\usepackage{graphicx}
\usepackage{multirow}
\usepackage{tabularx}
\usepackage{bm}
\usepackage{booktabs}
\usepackage{subcaption}
\usepackage[colorinlistoftodos,prependcaption,textsize=tiny]{todonotes}

\usepackage{gb4e}
\noautomath

\newcommand{\fem}[1]{\textcolor{cyan}{#1}}
\newcommand{\masc}[1]{\textit{\textcolor{purple}{#1}}}

\usepackage[T1,T2A]{fontenc}
\usepackage[utf8]{inputenc}
\usepackage[english,russian]{babel}
\usepackage{tempora}

\title{Automatically Identifying Gender Issues in\\ Machine Translation using Perturbations}

\author{Hila Gonen \\
  Bar-Ilan University \\
  \texttt{hilagnn@gmail.com} \\\And
  Kellie Webster \\
  Google Research NYC \\
  \texttt{websterk@google.com} \\}

\date{}

\begin{document}
\maketitle
\begin{abstract}

The successful application of neural methods to machine translation has realized huge quality advances for the community.
With these improvements, many have noted outstanding challenges, including the modeling and treatment of gendered language.
While previous studies have identified issues using synthetic examples, we develop a novel technique to mine examples from real world data to explore challenges for deployed systems.
We use our method to compile an evaluation benchmark spanning examples for four languages from three language families, which we publicly release to facilitate research.
The examples in our benchmark expose where model representations are gendered, and the unintended consequences these gendered representations can have in downstream application.

\end{abstract}

\section{Introduction}

Machine translation (MT) has realized huge improvements in quality from the successful application and development of neural methods \cite{kalchbrenner-blunsom-2013-recurrent-continuous,cho-etal-2014-learning,NIPS2017_7181,johnson-etal-2017-googles,chen-etal-2018-best}.
As the community has explored this enhanced performance, many have noted the outstanding challenge of modeling and handling gendered language \cite{K18,EJM19}.
We extend this line of work, which identifies issues using synthetic examples manually curated for a target language \cite{SSZ19,CKK19}, by analyzing real world text across a range of languages to understand challenges for deployed systems.

In this paper, we explore the class of issues which surface when a neutral reference to a person is translated to a gendered form (e.g. in Table~\ref{example}, where the English \emph{counselor} and \emph{nurse} are translated into the French \emph{conseiller} (masculine) and \emph{infirmi\`ere} (feminine).  
For this class of examples, the MT task requires a system to produce a single translation without source cues, thus exposing a model's preferred gender for the reference form.

With this scope, we make two key contributions.
First, we design and implement an automatic pipeline for detecting examples of our class of gender issues in real world input, using a BERT-based perturbation method novel to this work.
A key advantage of our pipeline beyond previous work is its extensibility: a) beyond word lists; b) to different language pairs and c) parts of speech. 
Second, using our new pipeline, we compile a dataset that we make publicly available to serve as a benchmark for future work.
We focus on English as the source language, and explore four target gendered languages across three language families (French, German, Spanish, and Russian).
Our examples expose where MT encodings are gendered, finding new issues not covered in previous manual approaches, and the unintended consequences of this for translation.

\begin{table*}[h!]
	
	\begin{center}
		
		\resizebox{\textwidth}{!}{
			\begin{tabular} { p{9cm} | p{9cm} | l   }
				
				Source Sentence (En) & Translation (Fr) & M/F\\ \hline \hline
				
				so is that going to affect my chances of becoming a \textbf{counselor}? & Alors, est - ce que cela va affecter mes chances de devenir \masc{conseiller}? & M \\ \hline
				
				so is that going to affect my chances of becoming a \textbf{nurse}? & Alors, est - ce que cela va affecter mes chances de devenir \fem{infirmi\`ere}? & F \\ \hline \hline

		\end{tabular}}
		\caption{An example from our dataset of a minimal pair of English gender-neutral source sentences, translated into two different genders in French. Red (italic) stands for masculine, cyan (normal) stands for feminine.} 
		\label{example}
		
	\end{center}
\end{table*}

\section{Gender Marking Languages}

Gender-marking languages have rich grammatical systems for expressing gender \cite{C91}.
To produce a valid sentence in a gender-marking language, gender may need to be marked not only on pronouns (\emph{he}, \emph{she}), as it is in English, but also nouns and even verbs, as well as words linked to these gendered nouns and verbs.
This means that translating from a language like English, with little gender marking, to a gender-marking language like Spanish, requires a system to produce gender markings that may not have explicit evidence in the source.
For instance, \emph{The tall teacher} from English could be translated into the Spanish \emph{\textbf{La} maestr\textbf{a} alt\textbf{a}} (feminine) or \emph{\textbf{El} maestr\textbf{o} alt\textbf{o}} (masculine).

\section{Automatic Detection of Gender Issues}

The class of issues we are interested in are those where translation to a gender-marking language exposes a model's gender preference for a personal reference.
The examples we find that demonstrate this are English sentence-pairs, a minimal pair differing by only a single word, e.g. \emph{doctor} being replaced by \emph{nurse}.
In each of our examples, this minimal perturbation does not change the gender of the source but gives rise to gender differences upon translation, e.g. \emph{doctor} becoming masculine and \emph{nurse} feminine.

In this section, we present a simple, extensible method to mine such examples from real-world text.
Our method does not require expansive manually-curated word lists for each target language, which enables us to discover new kinds of entities that are susceptible to model bias but are not usually thought of this way.
Indeed, while we demonstrate its utility with nouns with four target languages, our method is naturally extensible to new language pairs and parts of speech with no change in design.

\paragraph{Filtering source sentences}

Our first step is to identify sentences that are gender neutral and that include a single human entity, e.g. \emph{A \underline{doctor} works in a hospital}.
We focus on human entities since these have been the target of previous studies and present the largest risk of gender issues in translation.

We use a BERT-based Named Entity Recognition (NER) model that identifies human entities, and exclude sentences that have more than one token tagged as such. 
We also remove sentences in which the entity is a gendered term in English\footnote{\url{https://github.com/tolga-b/debiaswe/blob/master/data/gender_specific_full.json}\\\url{https://github.com/uclanlp/gn_glove/tree/master/wordlist}} (e.g. \textit{mother, nephew}), a name, or not a noun.

Note that all the sentences we get are naturally occurring sentences, and that we do not use any templates or predefined lists of target words that we want to handle.

\paragraph{Perturbations using BERT}
We use BERT as a masked language model to find words which can substitute for the human entity identified in the previous filtering step, e.g. \emph{doctor} $\rightarrow$ \emph{nurse}.
We aim to get natural-sounding output and maintain extensibility, and thus do not use predefined substitutions.
We cap our search to the first 100 candidates BERT returns, accepting the first 10 which are tagged as person, and for which the resulting sentences also pass the filtering step. 

\paragraph{Translation}
We translate each of the generated sentences into our target languages using Google Translate\footnote{\url{https://translate.google.com/}}.
\emph{A \underline{doctor/nurse} works in a hospital} $\rightarrow$ \emph{\underline{Un doctor/Una enfermera} trabaja en un hospital}.

\paragraph{Alignment}

We align tokens in the original and translated sentences using fast-align \cite{DLG10}.
This is needed in order to know which token in the translation output is the focus entity in the source sentence, whose gender we want to analyze.

\paragraph{Gender Identification}
We use a morphological analyzer, implemented following \citet{KAA17}, to tag the gender of the target word.

\paragraph{Identifying Examples}

The final step of our pipeline is identifying pairs of sentences to include in our dataset, pairs where different genders are assigned to the human entity.
Our example would be included since \emph{doctor} is translated with the masculine form \emph{Un doctor} while \emph{nurse} is translated with the feminine form \emph{Una enfermera}.

\section{Challenge Dataset}

We compile our final dataset from the output of this pipeline, and explore its properties to understand the issues it represents for deployed systems.

\subsection{Random Sampling}

In our final dataset, we include both examples that passed the final example identification step above (pairs referred to as ``at risk''), as well as a random selection that did not (``not at risk'').
We do this in order to not be constrained too heavily by our choice of translation model; 
if we did not, we would have no chance of inspecting examples that our system did not spot as at risk but other models might have.

\begin{table*}[h!]
	
	\centering
		
		\resizebox{\textwidth}{!}{
			\begin{tabular} {l| p{8.2cm} | p{8.2cm} |  p{0.5cm}   }
				
				 & Source Sentence (En) & Translation & M/F\\ \hline \hline
				
				\multirow{6}{*}{Fr} & also should i ask the \textbf{manager} what the pay would be if i got the job prior to flying out? &  De plus, devrais - je demander au \masc{gestionnaire} quel serait le salaire si je obtenais le poste avant de prendre l'avion? & M \\ 
				& also should i ask the \textbf{secretary} what the pay would be if i got the job prior to flying out? & De plus, devrais - je demander à la \fem{secrétaire} quel serait le salaire si je obtenais le poste avant mon départ? & F \\ \hline \hline
				
				\multirow{4}{*}{De}  & currently thinking about learning a trade (mostly a \textbf{electrician}). & Derzeit über das Erlernen eines Gewerbes nachdenken (meistens \masc{Elektriker}).  & M \\ 
				& currently thinking about learning a trade (mostly a \textbf{cook}). & Derzeit über das Erlernen eines Gewerbes nachdenken (meistens eine \fem{Köchin}). & F \\  \hline \hline

				\multirow{6}{*}{Es}  & - decided to become a \textbf{teacher}: spent a year working 2 jobs and doing prerequisites for a masters in education. &  - Decidí ser \fem{maestra}: pasé un año trabajando en 2 trabajos y haciendo requisitos previos para una maestría en educación. & F  \\ 
				& - decided to become a \textbf{lecturer} : spent a year working 2 jobs and doing prerequisites for a masters in education. & - Decidí ser \masc{profesor}: pasé un año trabajando en 2 trabajos y haciendo requisitos previos para una maestría en educación. &  M\\  \hline \hline
				
                \multirow{2}{*}{Ru}  & i read about a \textbf{psychologist} who upgraded into becoming a m.d. & Я читал о \masc{психологе}, который превратился в Md. &  M \\
                & i read about a \textbf{nurse} who upgraded into becoming a m.d. & Я читал о \fem{медсестре}, которая превратилась в доктора медицины. & F \\ \hline \hline

		\end{tabular}}
		\caption{Examples from our dataset of a minimal pair of English gender-neutral source sentences, translated into two different genders in all target languages. Red (italic) stands for masculine, cyan (normal) stands for feminine.} 
	\label{tab:examples}
\end{table*}

\subsection{Fixed Grammatical Gender Rating}

When we inspected the examples identified as at risk by our pipeline, the major source of error we found pertained to the issue of fixed grammatical gender.
Consider the example in Figure~\ref{example_gg}:

\begin{figure}[h!]
	\scalebox{0.75}{
	\fbox{\parbox{9.8cm}{			
			
			Sentence 1:

			\hspace{0.2cm} En: you don't have to be the \textbf{victim} in whatever.

			\hspace{0.2cm} Fr: vous ne devez pas \^etre la \fem{victime} de quoi que ce soit.

			Sentence 2:

			\hspace{0.2cm} En: you don't have to be the \textbf{expert} in whatever.
			
			\hspace{0.2cm} Fr: vous ne devez pas \^etre \masc{l'expert} en quoi que ce soit.
	}}}
	
	\caption{An example from our dataset, with fixed grammatical gender. Red (italic) stands for masculine, cyan (normal) stands for feminine.}
	\label{example_gg}
	
\end{figure}
	
In this example, the word \emph{victim} in the first English sentence is identified by our tagger as a human entity. 
However, its French translation \emph{victime} is feminine by definition, and cannot be assigned another gender regardless of the context, causing a false positive result.

We attempted to filter these examples automatically but came across a number of challenges.
Most critically, we found no high-quality, comprehensive dictionary that included the required information for all languages, and heuristics we applied were noisy and not reliable.\footnote{We tried both using a morphological lexicon and a predefined word list in English. Both methods performed poorly, filtering too many or too few sentence pairs, respectively.}
We observed that the underlying reason for these challenges was that there is no closed list of grammatically-fixed words as languages are evolving to be more gender-inclusive.
In order to maximize and guarantee data quality, and to be sensitive to the nuances of language change, we decided to add a manual filtering step after our pipeline to select the positive (at risk) examples.

We note that the problem of fixed grammatical gender is particular to nouns.
Our pipeline is naturally extensible across parts of speech and we would not expect the same issues in future work perturbing adjectives or verbs.

\subsection{Dataset Statistics}

To create our dataset we mine text from the subreddit ``career''.\footnote{\url{https://www.reddit.com/r/Career/}}
From 29,330 sentences, we found 4,016 which referred to a single, non-gendered human entity.
Introducing perturbations with BERT into these 4,016 sentences yielded 40,160 pairs. 
Out of those, 592 to 1,012 pairs are identified as at risk by our pipeline, depending on the target language. 
We asked humans to manually identify 100 true at risk examples for the final dataset, which was achieved for all languages except Russian, where we have 59 pairs.\footnote{Out of the pairs identified using our pipeline, between 1/3 - 1/10 were selected by the annotators, depending on the language.}
To this 100, we add a further 100 randomly sampled negative examples for each language.
Table~\ref{tab:examples} shows a representative example for each language-pair.

\subsection{Exploratory Analysis}

Table~\ref{top5} lists the most frequent focus personal references in each language-pair among the positive (at risk) and negative (not at risk) examples, along with the ratio between times the reference form was translated as masculine compared to feminine.
Words with extreme values of this ratio indicate cases where a model has a systematic preference for one gender over another, i.e. a gendered representation.

Among the negative examples, we see a prior for  masculine translations across all terms.
Positive examples break from this prior by exposing reference forms with a feminine preference:
\emph{nurse} and \emph{secretary} are the most consistently feminine forms, consistent with the Bureau of Labor statistics used in previous work \cite{caliskan2017semantics}.

\begin{table}
	
	\begin{center}
		\scalebox{0.88}{
			\begin{tabular} { l | l | r || l | r  }
				
				 & Positive & M:F & Negative & M:F  \\ \hline \hline
			
				\multirow{5}{*}{Fr} & nurse & 0:36 & manager & 685:1 \\ 
				
				& secretary & 0:17 & employee & 406:0 \\ 
				
				& teacher & 7:1 & employees & 364:0 \\ 
				
				& assistant & 1:7 & parents & 353:0 \\ 
				
				& manager & 8:0 & teacher & 337:0 \\  \hline \hline
				\multirow{5}{*}{De} & secretary & 0:27 & manager & 594:0 \\ 
				
				& nurse & 0:21 & employees & 409:1\\ 
				
				& teacher & 3:7 & friends& 359:0\\ 
				
				& receptionist & 0:9 & employee & 320:0\\ 
				
				& manager & 7:0 & students & 316:0\\ 	\hline \hline
				\multirow{5}{*}{Es} & teacher & 4:29 & manager& 691:0 \\ 
				
				& nurse &  0:31 & employee & 446:0\\ 
				
				& secretary & 0:26 & friends & 380:0 \\ 
				
				& writer & 8:0 &  parents & 374:0 \\ 
				
				& employee & 5:0 & supervisor & 345:0 \\ 	\hline \hline
				\multirow{5}{*}{Ru} & nurse & 0:32 & manager & 713:0\\ 
				
				& babysitter & 0:13 & employees & 519:0 \\ 
				
				& nurses & 0:5 & friends & 439:0 \\ 
				
				& dishwasher & 0:4 & students & 417:0 \\ 
				
				& technician & 3:0 & employee & 392:0 \\  \hline \hline
			
		\end{tabular}}
		\caption{Top five human reference forms in our dataset, and their ratio of times they are translated as masculine compared to feminine. Positive indicates that the examples were taken from the at-risk group from our pipeline, and negative from the random sample among the not at-risk group.} 
		\label{top5}
		
	\end{center}
\end{table}

Figure~\ref{example_all} shows two sentence pairs that appear as positive examples across all four language-pairs.
Two of the three forms, \emph{nurse} and \emph{mechanic}, are consistent with the gender statistics of \citeauthor{caliskan2017semantics};
the association of \emph{fighter} with the masculine gender is a new discovery of our method.

\begin{figure}
	\scalebox{0.75}{
	\fbox{\parbox{9.8cm}{			
			
			Sentence pair 1:

			\hspace{0.2cm} Original: you need to have experience working with hydraulic lifts, \& they like to see that you've worked or trained as a \textbf{mechanic}. 

			\hspace{0.2cm} Substitution: you need to have experience working with hydraulic lifts, \& they like to see that you've worked or trained as a \textbf{nurse}.\\

			Sentence pair 2:

			\hspace{0.2cm} Original: in fact, probably not even as a seasoned \textbf{nurse}. 

			\hspace{0.2cm} Substitution: in fact, probably not even as a seasoned \textbf{fighter}.
			
	}}}
	
	\caption{Two sentence pairs from our dataset that found to be shared between all four target languages.}
	\label{example_all}
	
\end{figure}

\section{Related Work}

Our study builds on the literature around gender bias in machine translation.
\citet{CKK19} use sentence templates to probe for differences in Korean pronouns.
\citet{PA19} and \citet{SSZ19} also use sentence templates, but filled with word lists, of professions and adjectives in the former, and professions in the latter.
A separate but related line of work focuses on generating correct inflections when translating to gender-marking languages \cite{VHW18,MAG19}.

\section{Conclusion}

The primary contribution of our work is a novel, automatic method for identifying gender issues in machine translation.
By performing BERT-based perturbations on naturally-occurring sentences, we are able to identify sentence pairs that behave differently upon translation to gender-marking languages.
We demonstrate our technique over human reference forms and discover new sources of risk beyond the word lists used previously.
Furthermore, the novelty of our approach is its natural extensibility to new language pairs, text genres, and different parts of speech.
We look forward to future work exploring such applications.

Using our new method, we compile a dataset across four languages from three language families.
By publicly releasing our dataset, we hope to enable the community to work together towards solutions that are inclusive and equitable to all.

\section*{Acknowledgements}

We thank Melvin Johnson for his helpful feedback throughout this project, Dan Garrette for helping with some parts of the pipeline, and Dani Mitropolsky, Vitaly Nikolaev and Marisa Rossmann for the help with filtering the dataset.

\bibliography{bib}
\bibliographystyle{acl_natbib}

\end{document}